\documentclass[twocolumn, letter]{article}
\usepackage{amsmath,amsfonts,graphicx}

\twocolumn
\usepackage{color,soul,bm}

\def\Zb{\textbf{Z}}

\def\mub{\boldsymbol{\mu}}
\def\sigb{\boldsymbol{\sigma}}

\def\xt{{\bf x}_t}

\def\xlt{{\bf x}_{< t}}
\def\xleT{{\bf x}_{\leq T}}
\def\hlt{{\bf h}_{t-1}}
\def\hnt{{\bf h}_{t}}
\def\xlet{{\bf x}_{\leq t}}
\def\psit{\boldsymbol{\psi}_t}
\def\qphi{q_{\boldsymbol{\phi}}}
\def\gphi{g_{\boldsymbol{\phi}}}
\def\hphi{h_{\boldsymbol{\phi}}}
\def\zt{{\bf z}_t}
\def\zlt{{\bf z}_{< t}}
\def\zlet{{\bf z}_{\leq t}}
\def\zleT{{\bf z}_{\leq T}}
\def\z{{\bf z}}
\def\KL{\mathrm{\bf KL}}
\def\ELBO{\mathrm{\bf ELBO}}
\def\log{\mathrm{log}}
\def\ex{\mathbb{E}}

\title{Semi-Implicit Stochastic Recurrent Neural Networks}
\author{%
  Ehsan Hajiramezanali$^{\dag}$\thanks{Both authors contributed equally.} , Arman Hasanzadeh$^{\dag *}$, Nick Duffield$^{\dag}$, Krishna Narayanan$^{\dag}$,\\ Mingyuan Zhou$^{\ddag}$, Xiaoning Qian$^{\dag}$
\\
\\
  $\dag$ Department of Electrical and Computer Engineering, Texas A\&M University\\
  \texttt{\{ehsanr, armanihm, duffieldng, krn, xqian\}@tamu.edu}\\
  $\ddag$ McCombs School of Business, University of Texas at Austin\\
  \texttt{mingyuan.zhou@mccombs.utexas.edu}
}
\date{}

\begin{document}
%
\maketitle
\begin{abstract}
Stochastic recurrent neural networks 
with latent random variables of complex dependency structures have shown to be more successful in modeling sequential data than deterministic deep models. However, the majority of existing methods have limited expressive power due to the Gaussian assumption 
of 
latent variables. In this paper, we advocate learning implicit latent representations using semi-implicit variational inference to further increase model flexibility.
Semi-implicit 
stochastic
recurrent neural network (SIS-RNN) is developed to enrich inferred model posteriors 
that may have no analytic density functions, as long as independent random samples can be generated via reparameterization. Extensive experiments in different tasks on real-world datasets show that SIS-RNN outperforms the existing methods.
\end{abstract}
%
%
%
\vspace{-0.05in} 
\section{Introduction}
\label{sec:intro}
\vspace{-0.1in} 

Deep auto-regressive models, such as recurrent neural networks (RNNs), are widely used for modeling sequential data due to their effective representation of long-term dependencies \cite{hasanzadeh2017psta,khorasgani2019fault,hajiramezanali2019scalable,hajiramezanali2018differential,hajiramezanali2012maneuvering1,hajiramezanali2012maneuvering,hajiramezanali2013stochastic}. It has been shown that inducing uncertainty in hidden states of deep auto-regressive models could drastically improve their performance in many applications such as speech modeling, text generation, sequential image modeling and dynamic graph representation learning~\cite{goyal2017z,hajiramezanali2019variational,chien2019variational,zhang2019improve,ruan2019condition,vyas2019learning,hajiramezanali2018bayesian,fouladi2013denoising,armandpour2019robust}. These methods integrate the variational auto-encoder~(VAE) framework 
with deep auto-regressive models
to infer stochastic latent variables, 
which can capture higher-level semantic abstraction (e.g. objects, speakers, or graph modules/communities) from the observed variables in a sequence (e.g. pixels, sound-waves, or partially observed dynamic graphs).

Existing stochastic recurrent models, while having different encoder and decoder structures, have restricted expressive power due to the commonly adopted Gaussian assumption on prior and posterior distributions of latent variables. The Gaussian assumption has a
well-known issue in underestimating the variance of the posterior \cite{blei2017variational}, which can be further amplified by mean field variational inference (MFVI).
This issue is often attributed to two key factors: 1) the mismatch between the restricted representation power of the variational family $Q$ and the complexity of the posterior to be approximated by $Q$; 2) the use of $\KL$ divergence, which is an asymmetric measure for the distance between $Q$ and the posterior \cite{yin2018semi,hasanzadeh2019semi}.

In this paper, we break the Gaussian assumption and propose a 
semi-implicit stochastic recurrent neural network (SIS-RNN)
that is capable of inferring implicit posteriors for sequential data while maintaining simple optimization. Inspired by semi-implicit variational inference (SIVI) \cite{yin2018semi}, we impose a semi-implicit hierarchical construction 
on a backbone RNN
to represent the posterior distribution of stochastic recurrent layers. SIVI 
enables a flexible (implicit) mixing distribution for variational inference of our proposed 
SIS-RNN. As a result, even if the marginal of the hierarchy is not tractable, its density can be evaluated by Monte Carlo estimation. Our proposed framework is capable of modeling skewness, kurtosis, multimodality, and other characteristics that are exhibited by the posterior of latent variables
but fail to be captured by the mean-field Gaussian variational family.
Our experiments demonstrate the superior performance of our proposed model in sequential image modeling and language modeling on multiple real-world datasets.

\vspace{-0.05in} 
\section{Preliminaries}
\vspace{-0.1in} 

\subsection{Semi-implicit variational inference (SIVI)} 
\vspace{-0.05in} 

SIVI has been proposed by \cite{yin2018semi} as a method for inferring implicit posteriors while maintaining simple
optimization.
SIVI assumes that the parameters of the posterior, $\psi$, are drawn from an implicit distribution instead of taking deterministic values. This hierarchical construction enables flexible mixture modeling and allows to have richer variational posteriors. More specifically, let $\Zb \sim q(\Zb \,|\, \psi)$ and $\psi \sim q_{\phi}(\psi)$, with 
$\phi$ denoting the distribution parameters to be inferred, and $q(\Zb \,|\, \psi)$ be the posterior distribution. 
Marginalizing $\psi$ out leads to the random variables $\Zb$ drawn from a distribution family $\boldsymbol{\mathcal{H}}$ indexed by variational parameters $\phi$, expressed as
\begin{equation}\label{equ : family}
    \boldsymbol{\mathcal{H}} =\left \{ \hphi(\Zb) \ : \ \hphi(\Zb) =  \int_{\psi}  q(\Zb \,|\, \psi) \qphi(\psi)\ d\psi \right \}.
\end{equation}

The essence of the semi-implicit formulation is that while the 
conditional posterior $q(\Zb \,|\, \psi)$ is explicit and analytic, the marginal distribution, $\hphi(\Zb)$ is often implicit. Note that, if $\qphi$ equals a delta function, then $\hphi$ is an explicit distribution. Unlike regular variational inference that assumes independent latent dimensions, SIVI does not impose such a constraint. This enables the resulting variational distributions to model very complex multivariate distributions such as multimodal or skewed distributions, which can not be captured by vanilla variational inference due to its often restricted exponential family assumption over both prior and posterior.

\vspace{-0.05in} 
\subsection{Variational recurrent neural network (VRNN)}
\vspace{-0.05in} 

VRNN \cite{chung2015recurrent} combines VAE with RNN to increase the expressive power of RNN and better model variability observed in highly structured sequential data. In VRNN, in addition to hidden states of RNN, a latent random variable is used to summarize past information. More specifically, given observations $\xlet$ and the stochastic variables $\zlet$, model likelihood
$p(\xt|\zlet,\xlt)$, and prior $p(\zt|\zlt,\xlt)$, 
we approximate the posterior $p(\zt | \xlet, \zlt)$ with a variational distribution $q(\zt | \psit)$ that is required to be explicit. We learn the variational parameters by minimizing $\KL(q(\zt | \psit) || p(\zt | \xlet, \zlt))$, the $\KL$ divergence of $p(\zt | \xlet, \zlt)$ and $q(\zt | \psit)$. Knowing that
\begin{equation}
    \log p({\bf x}_{\leq T}) = \ELBO + \sum_{t=1}^T\KL(q(\zt | \psit) || p(\zt | \xlet, \zlt)), \nonumber
\end{equation}
with $\ELBO=$
\begin{equation}
\label{eq:elbo}
    - \mathbb{E}_{q({\bf z}_{\le T} | {\bf x}_{\le T})} \left [ \sum_{t=1}^T \left( \log q(\zt | \psit) - \log p(\xlet, \zlet) \right) \right ], 
\end{equation}
minimizing $\KL(q(\zt | \psit)\, ||\, p(\zt | \xlet, \zlt))$ is hence equivalent to maximizing the $\ELBO$ \cite{chung2015recurrent}. Note that past information, i.e. $\xlet$, is transformed through RNN hidden states $\mathbf{h}_t$ as detailed below.

\vspace{-0.05in} 
\section{SIS-RNN}
\vspace{-0.1in} 
We introduce our SIS-RNN that imposes a distribution over parameters of posterior in VRNN, i.e. $\psit \sim q(\psit)$, instead of simply taking deterministic parameters. 

\vspace{0.25cm}
\noindent{\bf Model construction.}
Assuming $\psit \sim \qphi(\psit | \xlet, \zlt)$, where $\bm{\phi}$ denotes the parameters of the distribution to be inferred, the semi-implicit variational distribution for $\zt$ can be defined in a hierarchical manner as 
\begin{equation}
\label{eq:si1}
    \zt \sim q(\zt | \psit), \qquad \psit \sim \qphi(\psit | \xlet, \zlt).
\end{equation}

We impose an auto-regressive model to capture long-term dependency in the mixing distribution by exploiting an RNN architecture, that runs through the sequence as follows:
\begin{equation}
\label{eq:rnn}
    \hnt = f_\theta(\xt, \zt, \hlt),
\end{equation}
where $f$ is a deterministic non-linear transition function, and $\theta$ is the parameter set of $f$ to infer. By coupling the observations and latent variables using the recurrence equation~(\ref{eq:rnn}), SIS-RNN in~(\ref{eq:si1}) can be equivalently expressed as
\begin{equation}
    \zt \sim q(\zt | \psit), \qquad \psit \sim \qphi(\psit | \xt, \hlt).
\end{equation}

Marginalizing $\psit$ out leads to the random variables $\zt$ drawn from the distribution family $\bm{\mathcal{G}}$ indexed by variational parameters $\bm{\phi}$, expressed as
\begin{equation}
    \bm{\mathcal{G}} =\left \{ \gphi(\zt | \xt, \hlt) =  \int_{\psit}  q(\zt | \psit) \qphi(\psit | \xt, \hlt) d\psit \right \}
\end{equation}
While the variational distribution $q(\zt | \psit)$ 
is required to be explicit,
there is no such a constraint on the mixing distribution $\qphi(\psit | \xt, \hlt)$ and it is only required to be reparameterizable. In addition, $q(\zt | \psit)$ 
can be reparameterizable, with $\zt \sim q(\zt | \psit)$ being generated by transforming random noise $\epsilon$ via $f(\epsilon, \psit)$
or allowing the $\ELBO$ in (\ref{eq:elbo}) to be analytic.
More specifically, SIS-RNN draws samples from $\qphi(\psit | \xt, \hlt)$ by transforming random noise $\epsilon_t$ via a deep neural network. Specifically, assuming that conditional posterior is Gaussian, then, $q(\zt | \psit)$ $\sim \mathcal{N}\left(\mub^{(t)}_{\text{encoder}}, \text{diag}((\sigb_{\text{encoder}}^{(t)})^2) \right)$, with $\left\{\mub^{(t)}_{\text{encoder}}, \sigb_{\text{encoder}}^{(t)}\right\} = \varphi^{\text{encoder}} (\zt, \hlt, \epsilon_t)$ where $\varphi^{\text{encoder}}$ is a neural network. This generally leads to an implicit distribution for $\qphi(\psit | \xt, \hlt)$ due to a non-invertible transform $\varphi^{\text{encoder}}$. Therefore, the marginal variational distribution $\gphi(\zt| \xt, \hlt) \in \bm{\mathcal{G}}$ is often implicit, unless $\qphi(\psit | \xt, \hlt)$ is conjugate to $q(\zt | \psit)$.

Note that if $\qphi(\psit | \xt, \hlt)$ degenerates to the delta function $\delta_{\psit^0}(\psit | \xt, \hlt)$, the semi-implicit variational family $\bm{\mathcal{G}}$ reduces to the original $\bm{\mathcal{Q}} = {q(\zt | \psit^0)}$ family, where $\bm{\mathcal{Q}} \subseteq \bm{\mathcal{G}}$, as discussed in \cite{chung2015recurrent}. Unlike MFVI that assumes independent latent variables $z_t^{(l)}$, this expansion significantly helps restore the dependencies between them if $\psi_{t}^{(l)}$
are not imposed to be independent of each other.
Under this construction, the temporal variational distribution can be factorized as
\begin{equation}
\label{eq:qfactor}
    q({\bf z}_{\le T} | {\bf x}_{\le T}) = \prod_{t=1}^T \gphi(\zt | \xlet, \zlt).
\end{equation}

Instead of imposing a standard multivariate Gaussian distribution with deterministic parameters, VAE in our SIS-RNN learns the prior distribution parameters based on the hidden states in
previous time steps. 
In particular, we can write the construction of the prior distribution adopted as follows,
\begin{equation}
\label{eq:prior}
\begin{aligned}
    \zt \sim \mathcal{N}\left(\mub^{(t)}_{\text{prior}}, \text{diag}((\sigb_{\text{prior}}^{(t)})^2) \right),
\end{aligned}
\end{equation}
where $\left\{\mub^{(t)}_{\text{prior}}, \sigb_{\text{prior}}^{(t)}\right\} = \varphi^{\text{prior}} (\hlt)$ denote the parameters of the conditional prior distribution. Therefore, 
the generative model can be
factorized as
\begin{equation}
\label{eq:pfac}
\begin{aligned}
    p(\xleT, \zleT) &= \prod_{t=1}^T p(\xt| \zlet, \xlt) p(\zt| \zlt, \xlt)\\
    &= \prod_{t=1}^T p(\xt| \zt, \hlt) p(\zt| \hlt),
\end{aligned}
\end{equation}
where the parameters of the generating distribution $p(\xt| \zt, \hlt)$ can be learned using neural networks $\varphi^{\text{decoder}}$. 

\vspace{0.25cm}
\noindent{\bf Learning.}
Since the parameters of the posterior are random variables, the ELBO
goes beyond the simple VRNN and using equations (\ref{eq:qfactor}) and (\ref{eq:pfac}), $\ELBO$ can be derived as follows:
\begin{equation}
\begin{aligned}
\mathcal{L} &= \sum_{t=1}^T \biggl\{ \ex_{\psit \sim \qphi(\psit | \xt, \hlt)} \ex_{\zt \sim q( \zt| \psit)} \log p(\xt| \zt, \hlt) \\
    &-  \KL \left(\ex_{\psit \sim \qphi(\psit | \xt, \hlt)} q(\zt | \psit)\, ||\, p(\zt| \hlt) \right) \biggr\}.
\end{aligned}
\end{equation}
Direct optimization of the ELBO is not tractable \cite{yin2018semi,hasanzadeh2019semi,hajiramezanali2019variational}. Hence to infer variational parameters of SI-VGRNN, we derive a lower bound for the ELBO as follows:
\begin{equation}
\begin{aligned}
    \mathcal{L} 
    &= \ex_{\z \sim q(\zleT | \xleT)} \left [ \log p(\xleT, \zleT) - \log q(\zleT | \xleT) \right ]\\
    & = \sum_{t=1}^T \ex_{\zt \sim \gphi(\zt | \xt, \hlt)} \log \frac{p(\xt| \zt, \hlt) p(\zt| \hlt)}{\gphi(\zt | \xt, \hlt)}\\
    &= \sum_{t=1}^T \biggr \{- \KL (\ex_{\psit \sim \qphi(\psit | \xt, \hlt)} q(\zt | \psit)\, ||\, p(\zt| \hlt)) \\
    &\,\,\,\,\,\,\,\, + \ex_{\psit \sim \qphi(\psit | \xt, \hlt)} \ex_{\zt \sim q( \zt| \psit)} \log p(\xt| \zt, \hlt) \biggl \} \\
    &\geq \sum_{t=1}^T \biggr\{- \ex_{\psit \sim \qphi(\psit | \xt, \hlt)} \KL (q(\zt | \psit)\, ||\, p(\zt| \hlt))\\
    &\,\,\,\,\,\,\,\, + \ex_{\psit \sim \qphi(\psit | \xt, \hlt)} \ex_{\zt \sim q( \zt| \psit)} \log p(\xt| \zt, \hlt) \biggl \} \\ 
    &= \sum_{t=1}^T \ex_{\psit \sim \qphi(\psit | \xt, \hlt)} \ex_{\zt \sim q( \zt| \psit)} \\
    &\qquad\qquad \log \left(\frac{p(\xt| \zt, \hlt) p(\zt| \hlt)}{q(\zt | \psit)}\right)=\underline{\mathcal{L}}.
\end{aligned}
\end{equation}
Note that we used the following inequality from \cite{yin2018semi} to derive $\underline{\mathcal{L}}$, $\ex_{\psit} \KL (q(\zt|\psit) || p(\zt)$ $\geq \KL(\ex_{\psit} q(\zt|\psit) || p(\zt))$ .

While Monte Carlo estimation of $\underline{\mathcal{L}}$ only requires $\qphi(\zt | \psit)$ to have an analytic density function and $\qphi(\psit | \xt, \hlt)$ to be convenient to sample from, $\gphi(\zt | \xt, \hlt)$ is often intractable, and so the Monte Carlo estimation of the
ELBO $\mathcal{L}$ is prohibited. Therefore, SIS-RNN evaluates the lower bound separately from the distribution sampling. While the combination of an explicit $\qphi(\zt | \psit)$ with an implicit $\qphi(\psit | \xt, \hlt)$ is as powerful as needed, it is computationally tractable.

As discussed in \cite{yin2018semi}, without early stopping optimization, $\qphi(\psit | \xt, \hlt)$ can converge to a point mass density, making SIS-RNN degenerated to vanilla VRNN. To avoid this problem, we impose a regularization term to the lower bound $\underline{\mathcal{L}}_K = \underline{\mathcal{L}} + B_K$ as inspired by SIVI \cite{yin2018semi}: 
\begin{equation*}
\begin{aligned}
    B_K = \sum_{t=1}^T &\ex_{\psit, \psit^{(1)}, 
    \dots, \psit^{(K)} \sim \qphi(\psit | \xt, \hlt)}\\
    &\qquad\qquad \KL (q(\zt | \psit)\, ||\, \Tilde{g}_{K}(\zt | \xt, \hlt)),
\end{aligned}
\end{equation*}
where $\Tilde{g}_{K}(\zt | \xt, \hlt)) =$
\begin{equation}\label{equ : lowbound}
\frac{\qphi(\psit | \xt, \hlt) + \sum_{k=1}^{K} \qphi(\psit^{(k)} | \xt, \hlt)}{K+1}.
\end{equation}
This leads to an asymptotically exact ELBO that satisfies $\underline{\mathcal{L}}_0 = \underline{\mathcal{L}}$ and
$\lim_{K\rightarrow \infty} \underline{\mathcal{L}}_K = \mathcal{L}$.

\vspace{-0.05in}
\section{Experiments}
\vspace{-0.1in}

\textbf{Sequential MNIST.} 
We first evaluate the performance of SIS-RNN in the task of sequentially generating pixels in MNIST digits, which is a common benchmarking test in evaluating sequence modeling methods. We consider the binarized MNIST dataset 
as in \cite{larochelle2011neural}. 
Following the previous works \cite{goyal2017z}, we used 60,000 samples for training and 10,000 for testing. A Gated Recurrent Unit (GRU) with one layer of 64 hidden units was the backbone RNN in SIS-RNN. Two 64-dimensional fully-connected layers were adopted to model $\varphi^{\text{prior}}$ in (\ref{eq:prior}). We used a neural network with three 128-dimensional fully-connected layers as $\varphi^{\text{encoder}}$ while injecting [150, 100, 50] dimensional Bernoulli noise.
The model was trained for 2000 epochs using the Adam optimizer with a 0.001 learning rate at the mini-batch size of 128. $K$ in (\ref{equ : lowbound}) gradually increased from 1 to 100 during the first 500 epochs and remained constant after that. 
We used cyclic annealing~\cite{liu2019cyclical} as the KL annealing 
in $\ELBO$ to gradually impose the prior regularization term and avoid posterior collapse. 
The performance of SIS-RNN and the comparison with
other methods are provided in Table~\ref{tab:mnist}. We  report exact  negative  log-likelihood (NLL), approximate NLL (with $\approx$ sign), or the variational lower bound (with $\leq$ sign) based on the competing methods. While 64 hidden units were chosen to have the same number of parameters as competing methods, we show increasing the number of hidden units to 128 significantly improves the performance of SIS-RNN without overfitting.

\begin{table}[t]
\caption{Comparison of the 
negative log-likelihood (NLL)
between various algorithms for sequential MNIST.}
\label{tab:mnist}
\vspace{-0.05in}
\begin{center}
\begin{small}
\begin{tabular}{lcr}
\hline
{\bf Model} & {\bf NLL} \\
\hline
DBN 2hl \cite{germain2015made}   & $\approx$84.55 \\
NADE \cite{uria2016neural}   & 88.33 \\
EoNADE-5 2hl \cite{raiko2014iterative}   & 84.68 \\
DLGM 8 \cite{salimans2015markov}   & $\approx$85.51 \\
DARN 1hl \cite{gregor2015draw}   & $\approx$84.13 \\
DRAW \cite{gregor2015draw}   & $\leq$80.97 \\
PixelVAE \cite{gulrajani2016pixelvae}   & $\approx$79.02 \\
P-Forcing\textsubscript{(3-layers)} \cite{lamb2016professor}   & 79.58 \\
PixelRNN\textsubscript{(1-layer)} \cite{van2016pixel}   & 80.75 \\
PixelRNN\textsubscript{(7-layers)} \cite{van2016pixel}   & 79.20 \\
MatNets \cite{bachman2016architecture}   & 78.50 \\
Z-Forcing\textsubscript{(1-layer)} \cite{goyal2017z}   & $\leq$ 80.60 \\
Z-Forcing\textsubscript{(1-layer)} + aux \cite{goyal2017z}   & $\leq$ 80.09 \\
TwinNet\textsubscript{(3-layers)}  \cite{serdyuk2018twin}   & $\leq$ 79.12 \\
VRNN\textsubscript{(1-layer)}   & $\leq$ 74.15 \\
\hline
{\bf SIS-RNN}\textsubscript{(1-layer)} 64 & {\bf 71.90} \\
{\bf SIS-RNN}\textsubscript{(1-layer)} 128 & {\bf 70.57} \\
\hline
\end{tabular}
\end{small}
\end{center}
\vskip -0.25in
\end{table}

\vspace{0.25cm}
\noindent\textbf{IAM-OnDB.} This human handwriting dataset contains 13,040 handwriting lines written by 500 writers \cite{liwicki2005iam}. The writing trajectories are represented as a sequence of $(x, y)$ coordinates together with binary indicators of pen-up/pen-down. We followed \cite{chung2015recurrent,lai2018stochastic} to preprocess and split the dataset. The experimental setup for IAM-OnDB is the same as that of the sequential MNIST experiment except that we used 256 hidden units for GRU to have the same number of parameters with competing methods. We report the average negative log-likelihood of test examples in Table~\ref{tab:iamondb}. For SIS-RNN, WaveNet, and RNN, we report the exact log-likelihood, while in the other cases, we
report the variational lower bound (with $\leq$ sign).
Our results show that SIS-RNN achieves higher log-likelihood, which supports our expectation that implicit latent random variables are helpful when modeling complex sequences.

\begin{table}[t]
\caption{Comparison of the average NLL between various algorithms for IAM-OnDB.}
\label{tab:iamondb}
\vspace{-0.05in}
\begin{center}
\begin{small}
\begin{sc}
\begin{tabular}{lcr}
\hline
{\bf Model} & {\bf Average NLL} \\
\hline
RNN \cite{chung2015recurrent}   & -1358 \\
VRNN \cite{chung2015recurrent}   & $\leq$-1384 \\
WaveNet \cite{lai2018stochastic}   & -1021 \\
SWaveNet \cite{lai2018stochastic}   & $\leq$-1301 \\
STCN \cite{lai2018stochastic}   & $\leq$-1338 \\
STCN-dense \cite{lai2018stochastic}   & $\leq$-1796 \\
\hline
{\bf SIS-RNN}\textsubscript{(1-layer)} & {\bf -1973} \\
\hline
\end{tabular}
\end{sc}
\end{small}
\end{center}
\vskip -0.25in
\end{table}

\vspace{0.3cm}
\noindent{\bf Language modeling.}
Due to their powerful model capacity by distribution-based latent representations, VAEs have become the generative models of choice for dealing with many natural language processing (NLP) tasks including language modeling \cite{bowman2015generating}.
This flexible representation allows capturing holistic properties of sentences, such as text style, topic, and high-level linguistic and semantic features. Generated samples from the prior latent distribution can further produce diverse and well-formed sentences through simple deterministic decoding.

Despite its popularity,  1) the adopted auto-regressive decoder, which is often implemented with an RNN, 
tends to ignore the latent variables in decoding, yielding ``posterior collapse'' \cite{bowman2015generating, liu2019cyclical}; 2) 
the Gaussian assumption imposed on the variational distribution restricts its variational inference capacity. 
While there exists a variety of methods to address the first problem by either changing the decoder \cite{yang2017improved} or applying KL annealing \cite{bowman2015generating,liu2019cyclical}, only a few works addressed the latter one including semi-amortized VAE (SA-VAE) \cite{kim2018semi}.  

It has been shown that having one latent variable for each sentence is more effective than including one latent variable for each word \cite{liu2019cyclical}. Therefore, for this experiment, we customized our SIS-RNN to have only one stochastic latent variable for each sequence of data, i.e. sentence.
More specifically, we only infer one variational latent variable from the last hidden state of RNN.
Moreover, we used a self-attention transformer as the decoder, i.e. $\varphi^{\text{decoder}}$, similar to \cite{hu2018texar}. 
We used the same experimental setting as in the previous works~\cite{bowman2015generating, yang2017improved,hu2018texar}. The rest of the hyper-parameters of our model are the same as those of our IAM-OnDB experiment. 
We consider two public datasets, the Yahoo \cite{yang2017improved} and
Penn Treebank~(PTB)~\cite{bowman2015generating}. While PTB is a relatively
small dataset with sentences of varying lengths,
Yahoo contains more samples with longer sentences. 
Table~\ref{tab:langmod} shows the perplexity~(PPL), sentence-level NLL and KL divergence of test samples. Not only SIS-RNN outperforms other methods in terms of NLL and PPL, but also checking KL values indicates that SIS-RNN does not suffer from posterior collapse. 
\textit{"how to stay in hot water when i get dizzy?"}, \textit{"i just like this girl and we have been friends for 4 yrs."}, and \textit{"i hate it personally."} are the generated examples from the model trained on Yahoo. 
\textit{"he probably showed it this month as a fundamental policy which includes the best of $<$unk$>$ and sales $<$EOS$>$"} and \textit{"the market 's bullish trend is underway $<$EOS$>$"} are two generated examples from the model trained on PTB.

\begin{table}[t]
\caption{Comparison of language modeling on two datasets.}
\label{tab:langmod}
\vspace{-0.05in}
\begin{center}
\resizebox{\columnwidth}{!}{%
\begin{small}
\begin{sc}
\begin{tabular}{c|ccccr}
\hline
{\bf Dataset} & {\bf Model} &  {\bf NLL}& {\bf PPL}& {\bf KL}
\\
\hline
 & VAE-LSTM \cite{bowman2015generating}  & 337.3 &  68.31 & 0.0\\
Yahoo & VAE-Transformer \cite{hu2018texar}  & 328.6 & 61.6 & 0.7 \\
& SA-VAE \cite{kim2018semi} & 327.2 &  60.1 & { 5.2}\\
& {\bf SIS-RNN} & {\bf 326.7} & {\bf 59.8} & 4.2\\
\hline
 & VAE-LSTM \cite{bowman2015generating}  &  102.1 & 105.2 & 0.0 \\
 & VAE-Transformer \cite{hu2018texar}  & 101.5 & 102.4  & 0.2\\
PTB & Cyclic-VAE \cite{liu2019cyclical} & 103.1 &  110.5 & {3.5}\\
& SA-VAE \cite{kim2018semi} & 102.6 & 107.1 & 1.2 \\
& {\bf SIS-RNN} & {\bf 101.2} & {\bf 101.8} & {1.6} \\
\hline
\end{tabular}
\end{sc}
\end{small}
}
\end{center}
\vskip -0.25in
\end{table}

\vspace{-0.05in}
\section{Conclusion}
\vspace{-0.05in}

We have proposed SIS-RNN, the first stochastic recurrent latent variable model with more expressive variational posteriors.
We argue that more flexible variational inference in SIS-RNN is a key to better modeling of the dependency in the sequential data.
We have tested SIS-RNN on three different tasks with SIS-RNN outperforming competing methods substantially.


\small
\bibliographystyle{IEEEbib}
\bibliography{refs}

\end{document}